\documentclass{article}

\usepackage[accepted]{icml2021}

\usepackage[utf8]{inputenc}
\usepackage[T1]{fontenc}
\usepackage{microtype}
\usepackage{dsfont}
\usepackage{bm}
\usepackage{graphicx}
\usepackage{amsmath,amsfonts}
\usepackage{booktabs}
\usepackage{multirow}
\usepackage{makecell}
\usepackage{textcomp}
\usepackage{array}
\usepackage{textcomp}
\usepackage{multirow}
\usepackage{float}
\usepackage{xspace}
\usepackage[export]{adjustbox}
\usepackage{nicefrac}
\usepackage{colortbl}
\usepackage{tablefootnote}
\usepackage[referable]{threeparttablex}
\usepackage{comment}
\usepackage{listings}
\usepackage{subcaption}
\usepackage{lipsum}

\usepackage{mathtools}

\DeclarePairedDelimiterX{\infdivx}[2]{(}{)}{%
	#1\;\delimsize\|\;#2%
}
\makeatletter
\@ifpackageloaded{icml\the\year}{
\usepackage{enumitem}
\setlist{nosep}}
{\usepackage[linesnumbered,ruled,vlined]{algorithm2e}
\SetKwRepeat{Do}{do}{while}
\DontPrintSemicolon

\SetCommentSty{commentsty}
}
\@ifclassloaded{acmart}{
\def\authornotetext#1{
	\g@addto@macro\@authornotes{%
	\stepcounter{footnote}\footnotetext{#1}}%
}}{
\usepackage[usenames,svgnames]{xcolor}
\usepackage{hyperref}
\hypersetup{
    colorlinks = true,
    breaklinks = true,
    bookmarksdepth = section,
    citecolor = DarkGreen,
    linkcolor = DarkRed,
    urlcolor = DarkBlue,
}}
\@ifclassloaded{llncs}{
\usepackage[square,comma,numbers,sort&compress]{natbib}
\bibliographystyle{splncsnat}

}{
\usepackage{amsthm}

\theoremstyle{remark}

}
\makeatother

\DeclarePairedDelimiter\abs{\lvert}{\rvert}%

\DeclarePairedDelimiter{\nint}\lfloor\rfloor

\DeclareMathAlphabet{\mathsfit}{\encodingdefault}{\sfdefault}{m}{sl}
\SetMathAlphabet{\mathsfit}{bold}{\encodingdefault}{\sfdefault}{bx}{n}

\usepackage[hang,flushmargin]{footmisc}
\usepackage{balance}
\usepackage{flushend}

\begin{document}
\twocolumn[
\icmltitle{Effective and Interpretable fMRI Analysis via \mbox{Functional Brain Network Generation}}
\icmlsetsymbol{correspondence}{\dag}

\begin{icmlauthorlist}
\icmlauthor{Xuan Kan}{cs}
\icmlauthor{Hejie Cui}{cs}
\icmlauthor{Ying Guo}{biostats}
\icmlauthor{Carl Yang}{cs,correspondence}
\end{icmlauthorlist}

\icmlaffiliation{cs}{Department of Computer Science, Emory University}
\icmlaffiliation{biostats}{Department of Biostatistics and Bioinformatics, Emory University}
\icmlcorrespondingauthor{Carl Yang}{j.carlyang@emory.edu}

\icmlkeywords{GCN, brain networks}

\vskip 0.3in
]

\printAffiliationsAndNotice{}

\begin{abstract}
Recent studies in neuroscience show great potential of functional brain networks constructed from fMRI data for popularity modeling and clinical predictions. However, existing functional brain networks are noisy and unaware of downstream prediction tasks, while also incompatible with recent powerful machine learning models of GNNs. In this work, we develop an end-to-end trainable pipeline to extract prominent fMRI features, generate brain networks, and make predictions with GNNs, all under the guidance of downstream prediction tasks. Preliminary experiments on the PNC fMRI data show the superior effectiveness and unique interpretability of our framework.
\end{abstract}


\section{Introduction}
In recent year, network-oriented analysis has become increasingly important in neuroimaging studies in order to understand human brain organizations in healthy as well as diseased individuals \citep{genderfunction, wangfilter, hierarchicalyin, brainnetworks, concepts}. There are abundant findings in neuroscience research showing that neural circuits are the key for understanding the differences in brain functioning between populations and the disruptions in neural circuits largely cause and define brain disorders \citep{insel2015brain,williams2016precision}. Functional magnetic resonance imaging (fMRI) is one of the most commonly used imaging modalities to investigate brain functions and organizations \citep{GANIS2002493,lindquist2008statistical,smith2012future}. There is a strong interest in neuroimaging community to predict clinical outcomes or classify individuals based on brain networks derived from fMRI images \citep{BrainNetCNN, predictsASD}.

Current network analysis typically takes the following approach \citep{modellingfmri, simpson2013analyzing,wangfilter}. First, functional brain networks are estimated based on individuals' fMRI data. This is usually done by selecting a brain atlas or a set of regions of interests and extracting fMRI blood-oxygen-level-dependent (BOLD) signal series from each node or brain regions. Then,  pairwise connectivity is calculated between node pairs using measures such as Pearson correlation and partial correlations. The calculated brain connectivity measures between all the node pairs are then used in the subsequent classification or prediction analyses to classify individuals or predict their clinical outcomes.
 However, the original BOLD signal series are often high-dimensional and noisy, and the brain networks constructed in this way are not customized towards specific downstream clinical predictions.

Recently, there is a growing trend in applying graph neural networks (GNNs) on brain connectivity matrices from fMRI data \citep{groupinn2019, Anirudh2019BootstrappingGC2019, Li2019GraphNN, brainGNN}. GNNs are the state-of-the-art deep learning methods on understanding graph-structured data which can combine graph structures and node features for various graph-related predictions \citep{nodeprediction, gin, infomax, linkprediction, yang2020heterogeneous}. Since the mechanism of most GNNs (i.e., message passing) is not compatible with existing functional brain networks which possess both positive and negative weighted edges but no proper node features, recent works also studied how to generate graphs or apply GNNs without pre-computed brain networks \citep{ Learninggraph2019, shang2021discrete, DBLP:conf/nips/YangZSL019}. Interestingly, \citep{huanghen2019} shows that applying GNNs on randomly generated graphs with proper node features can also provide prediction power, but such random graphs do not capture useful brain connectivity and do not support interpretable clinical analysis and prediction. 

\begin{figure*}[h]
    \centering
    \includegraphics[width=1.0\linewidth]{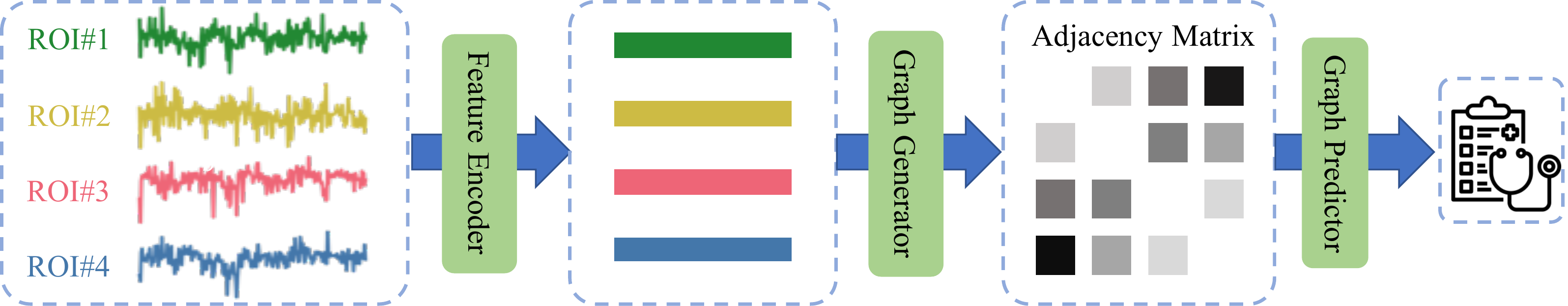}
    \caption{Overall framework of our end-to-end fMRI analysis pipeline with functional brain network generation. 
    }
    \label{fig:model}
\end{figure*}

In this work, in order to unleash the power of GNNs in network-based fMRI analysis while providing valuable interpretability regarding brain region connectivity, we propose to generate functional brain networks that are compatible with GNNs and customized towards downstream clinical predictions from fMRI data. Specifically, we develop an end-to-end differentiable pipeline from BOLD signal series to clinical predictions, which includes a feature extractor for denoising and reducing the dimension of raw time-series data, a graph generator for generating individual brain networks from the extracted features, and a graph predictor of GNN for clinical predictions from the generated brain networks (c.f.~Figure \ref{fig:model}).

We conduct preliminary experiments on the real-world fMRI dataset of PNC and focus on the downstream task of gender prediction. Our significant performance improvements over state-of-the-art time-series models and GNNs on existing functional brain networks demonstrate the plausible effectiveness of our method, whereas our in-depth analysis with visualizations on the generated brain networks showcase the unique interpretability of our method.



\section{Proposed Framework}
\subsection{Overview}
In this section, we elaborate the design of our end-to-end pipeline and its three main components as shown in Figure \ref{fig:model}.
Specifically, the input $\bm X\in \mathbb{R}^{n \times v \times t}$ denotes the BOLD time-series for regions of interest (ROIs) as the input, where $n$ is the sample size, $v$ is the number of ROIs and $t$ is the length of time-series. Each $\bm x\in \mathbb{R}^{v \times t}$ represents a sample (individual). 
The target output is the prediction label $\bm Y \in \mathbb{R}^{n \times |\mathcal{C}|}$, where $\mathcal{C}$ is the class set of $Y$ and $|\mathcal{C}|$ is the number of classes. 
As an intermediate output of the pipeline, we also generate a functional brain network $\bm A\in\mathbb{R}^{v\times v}$ for each sample $\bm x\in\mathbb{R}^{v\times t}$, which represents the brain connectivity matrix between ROIs.

\subsection{Feature Encoder}
\label{sec:fe}

In the feature encoder component $\operatorname{EXT}$, when $\operatorname{EXT}$ is set as bi-GRU with window size $\tau$, the process of generating $\bm{h}_{e} \in  \mathbb{R}^{v \times o}$ can be decomposed as two steps, where $o$ is the feature size of each ROI,
$$
\bm{h}_r = \operatorname{biRNN}([\bm x^{(z\tau-\tau):z\tau}]),  \bm{h}_r \in \mathbb{R}^{v \times 2\tau},
$$
$$
z=1,\cdots,\nint{\frac{t}{\tau}},
$$
$$
\bm{h}_{e} = \operatorname{MLP}(\bm{h}_r),
$$

Similarly, when $\operatorname{EXT}$ is set as $u$-layer 1D-CNN, the process can be decomposed like
$$
\bm{h}^u = \operatorname{CONV}_u(\bm{h}^{u-1}),
$$
$$
\bm{h}_{e} = \operatorname{MLP}(\operatorname{MAXPOOL}(\bm{h}^u)),
$$
where $\bm{h}^{0}= \bm x$ is the original data sample.

\subsection{Graph Generator}
In the middle stage between encoder and predictor, a feature-based and prediction-driven functional brain network can be generated from $\bm{h}_{e}$, formulated as the connectivity matrix $\bm A$, which
has ROIs along rows and columns and stores the pair-wise connectivity strengths. Unlike the basic (existing) approach for functional brain network construction that calculates the pairwise Pearson correlations between raw time-series of ROIs in $\bm x$ \citep{modellingfmri}, we generate $\bm A$ based on feature extracted from its time-series data as 
$$
\bm{h}_{A} = \operatorname{softmax}(\bm{h}_{e}),
$$
\begin{equation}
\label{equ:adj}
\bm{A} = \bm{h}_{A}\bm{h}_{A}^T,
\end{equation}
which is end-to-end learnable based on both raw time-series features and downstream prediction tasks. The softmax operation highlights the strong ROI connections, which generates skewed positive edge weights that are compatible with GNNs and valuable for interpretation. 

In order to facilitate the learning of brain networks beyond the sheer supervision of graph-based classification, we further apply three group-based and structure-based constraints during training, named as group inner loss, group intra loss and sparsity loss, respectively. Given a class $c \in \mathcal{C}$, $\mathcal{S}^{c}=\{i \mid Y_{i,c} = 1\}$ is the set containing all sample's index whose label is $c$. In order to construct group-based regularizers, we need to first obtain $\mu_c$ and $\sigma^2_c$ as the mean and variance of all samples' $\bm A$ with label $c$,
$$
\label{equ:mu}
\mu_c = \sum_{k \in \mathcal{S}^{c}}{\frac{\bm{A}^k}{\abs{\mathcal{S}^{c}}}}, 
$$
$$
\sigma^2_c = \sum_{k \in \mathcal{S}^{c}}{\frac{\left\|\bm{A}^k-\mu_c\right\|_{2}^{2}}{\abs{\mathcal{S}^{c}}}}.
$$
\textbf{Group Inner Loss} aims to minimize the difference of connectivity matrices within a class. Through the following derivation, this loss can be effectively calculated in $\mathcal{O}(n)$ time as
\begin{align*}
\label{equ:inner}
\begin{split}
L_{inner} & = \sum_{c \in \mathcal{C}} \sum_{i \in \mathcal{S}^{c}}{\frac{\left\|\bm{A}^i-\mu_c\right\|_{2}^{2}}{\abs{\mathcal{S}^{c}}}} \\
&= \sum_{c \in \mathcal{C}}{\sigma^2_c }.
\end{split}
\end{align*}
\textbf{Group Intra Loss} aims to maximize the difference of connectivity matrices across different classes, while keeping those within the same class similar. With proper derivation, this loss can also be calculated in $\mathcal{O}(n)$ time as
\begin{align*}
L_{intra} & =\sum_{a,b \in \mathcal{C}}{(\sigma^2_a +\sigma^2_b-\frac{\sum_{i \in \mathcal{S}^{a}} \sum_{j \in \mathcal{S}^{b}}\left\|\bm A^{i}-\bm A^{j}\right\|_{2}^{2}}{\abs{\mathcal{S}^{a}}\abs{\mathcal{S}^{b}}}}) \\
& = -\sum_{a,b \in \mathcal{C}} \left\|\mu_{a}-\mu_{b}\right\|_{2}^{2}.
\end{align*}
\textbf{Sparsity Loss} aims to enforce the sparsity of brain networks, so as to highlight the task-specific ROI connections, which is formulated as 
$$
L_{sparsity} =\frac{1}{nvv}\sum^n_{i=1}\left\|\operatorname{vec}(\bm A^i)\right\|_1.
$$

\subsection{Graph Predictor}  
With the initial node features $\bm F \in \mathbb{R}^{v \times f}$ of ROIs and the learnable connectivity matrix $\bm A$ from graph generator, we can apply a $k$-layer graph convolutional network \citep{nodeprediction} to compute the node embeddings through
$$
\bm{h}^{k} = \operatorname{ReLU}\left(\bm{A} \bm{h}^{k-1} W^{k}\right),
$$
where $W^{k}$ represents learnable parameters in convolutional layers and $\bm{h}^{0}=\bm F$. 
The graph-level embedding is obtained by summing up all the node embeddings.
A $\operatorname{BatchNorm1D}$ is the applied to avoid extreme large values. Finally, another $\operatorname{MLP}$ function is employed for prediction,
$$
\hat{y} =  \operatorname{MLP}\left(\operatorname{BatchNorm1D}\left(\sum_{p = 1}^{v} \bm{h}^{k}_p\right)\right).
$$
The supervised objective for prediction is the cross-entropy loss $L_{ce}$. Overall, our total loss function is composed of four terms: 
$$
L=L_{ce}+\alpha L_{inner}+\beta L_{intra}+ \gamma L_{sparsity},
$$
where $\alpha, \beta, \gamma$ are tunable hyper-parameters representing the weights of different regularizers.





\section{Experiments}
In this section, we evaluate the effectiveness and interpretability of our proposed framework. 

Specifically, the purpose of empirical studies on the effectiveness is to answer the following three questions.
\begin{itemize}
\item \textbf{RQ1.} How does our proposed framework outperform other baseline methods that directly model the time-series features?
\item \textbf{RQ2.} Do the graphs generated from our framework better facilitate GNNs compared with existing functional brain networks?
\item \textbf{RQ3.} What influences do the regularizers bring to the performance of our framework?
\end{itemize}

For interpretability study, our purpose is to investigate the advantages of our learnable graph and the consistency of its patterns with existing neuroscience discoveries. 

\begin{figure*}[h]
    \centering
    \includegraphics[width=1.0\linewidth]{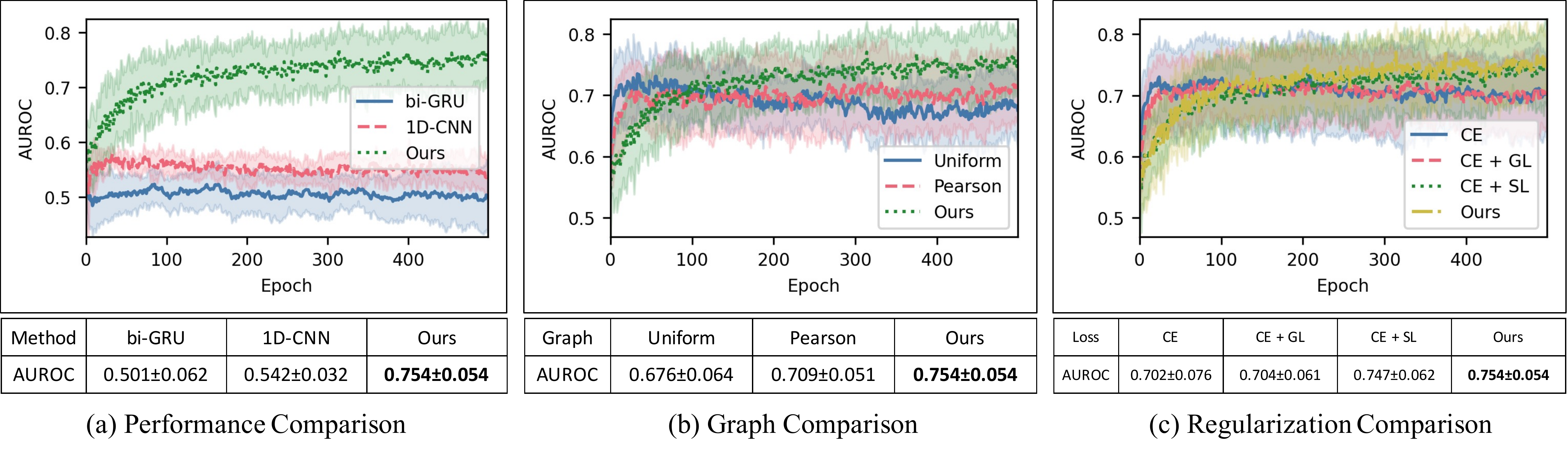}
    \caption{Performances of compared methods. All of the reported scores are evaluated on the testing set and are based on 5-fold cross-validation with 3 repetitions. The results shown in the tables are based on the models trained after 500 epochs. (a) Performance comparison with other Time-series model shows the superiority of intermediate brain network generation. (b) Graph comparison shows that our learnable graph is more compatible with GNN and can achieve improved performance than the two others with a large gap. (c) Ablation study of each loss component: GL represents Group Loss, SL represents Sparsity Loss and CE is Cross Entropy Loss.} 
    \label{fig:result}
\end{figure*}

\subsection{Experimental Settings}

\textbf{Dataset.} In this paper, we use the PNC dataset \citep{pnc} for the evaluation. PNC is a collaborative project from the Brain Behavior Laboratory at the University of Pennsylvania and the Children's Hospital of Philadelphia. It includes a population-based sample of individuals aged 8–21 years. 
Prior to analysis, we perform quality control on the rs-fMRI data such as displacement analysis to remove images with excessive motion \citep{genderfunction,wangfilter}. 503 subjects participants' rs-fMRI and DTI data met the quality control criterion and are used in our following analysis. Among these subjects, 289 (57.46\%) are female and the mean age is 15.28 years (SD = 3.11), which indicates that our dataset is balanced across genders. In our paper, we adapt the 264-node cortical parcellation system defined by \citep{power264} for connectivity analysis. The nodes are grouped into 10 functional modules that correspond to the major resting state networks \citep{10module}.  Standard pre-processing procedures are applied to the rs-fMRI data. For rs-fMRI, the preprocessing include despiking, slice timing correction, motion correction, registration to MNI 2mm standard space, normalization to percent signal change, removal of linear trend, regressing out CSF, WM, and 6 movement parameters, bandpass filtering (0.009–0.08), and spatial smoothing with a 6mm FWHM Gaussian kernel. Finally, each sample contains 264 nodes and is collected through 120 time steps.


\textbf{Metrics.} 
We perform gender prediction as the evaluation task. Since it is a binary classification problem and our dataset is balanced between classes, AUROC is the most comprehensive performance metric and is adopted here for performance comparison.


\textbf{Implement Details.} The weights of each loss component $\alpha, \beta, \gamma$ are respectively set as $10^{-3}$, $10^{-3}$ and $10^{-4}$. For the feature encoder, when 1D-CNN is adopted, the number of CNN layer $u$ is 3 and the kernel size is 16, with the kernel size of $\operatorname{MAXPOOL}$ setting as 4. When GRU is adopted as the feature encoder, the window size $\tau$ is 16 and the layer number of the GRU is 3. The feature dimension $o$ of $\bm{h}_e$ is set as 8. For the graph predictor, we set its layer number as 3 and construct the node feature $\bm F_p$ for the node $p$ as a vector of Pearson correlation scores between its time series data with those of all nodes contained in the graph. We use the Adam optimizer with the initial learning rate setting as $10^{-4}$ and a weight decay of $10^{-4}$ to train the model. The batch size is set as 16.




\subsection{Comparison with time-series models (RQ1)}
We compare our model with two commonly used models, 1D-CNN and bi-GRU, to handle temporal sequence. 1D-CNN and bi-GRU are both applied to encode BOLD time-series, then use a multilayer perceptron to directly make predictions based on the encoded features without building brain networks. 
The overall performance is presented in Figure \ref{fig:result}(a).
To ensure fairness, all hyper-parameters are shared across two baselines and our model. 
It can be seen that our model outperforms both baseline by significant margins (up to 20\% absolute improvements), which demonstrates the necessity of intermediate brain network generation. 
The comparison between two baselines also provide insights on the choice of feature encoder. As is shown by the curves, 1D-CNN consistently shows a better performance than bi-GRU. Therefore, 1D-CNN is adopted as the feature extractor in our end-to-end framework. 

\subsection{Comparison with other graphs (RQ2)}
We compare our learnable graphs $\bm{A}$ calculated from Equation \ref{equ:adj} with existing brain networks \citep{modellingfmri} and uniform graphs.
For existing brain networks $\mathbf{A}^P$, each entry is calculated as the the Pearson correlation coefficient between two raw time-series,
$$
\label{equ:perason}
\mathbf{A}_{p,q}^{P} = \abs{\operatorname{cov}(\bm x_p, \bm x_q)},
$$
where $\operatorname{cov}$ represents the co-variance function.
To isolate the influence of node features, we also use the uniform graphs $\mathbf{A}^{U}$ as a control group, which correspond to adjacency matrixes with all 1's,
$$
\mathbf{A}^{U} = \{1\}^{v \times v}.
$$
All these three types of graphs are paired with the same node features $\bm F$ and then processed by the same graph predictor of GNNs for the task prediction. 
As is shown in Figure \ref{fig:result}(b), GNNs with our learnable graph gains more improvements compared with other graphs, indicating that our learnable graph is more informative and compatible with GNNs. 

\subsection{Comparison with different model ablations (RQ3)}
In this section, we further verify the influence of two major regularizers in the graph generator: the Group Loss (GL) (including both Group Inner Loss and Group Intra Loss) and the Sparsity Loss (SL). Starting from the CE loss alone, we add one regularizer each time and see how it influences the performance compared with the model. The ablation results are shown in Figure \ref{fig:result}(c). We observe that the full model improves more stably than its three downgraded versions and finally achieves the highest performance. Specifically, CE+SL achieves close-to-optimal performance, demonstrating the effectiveness of sparsity constraints in generating informative brain networks. The smaller AUROC standard deviation of CE+CL compared with CE indicates that incorporating group losses can obtain more stable performance.

\subsection{Visualization}
\begin{figure}[]
    \centering
    \includegraphics[width=1.0\linewidth]{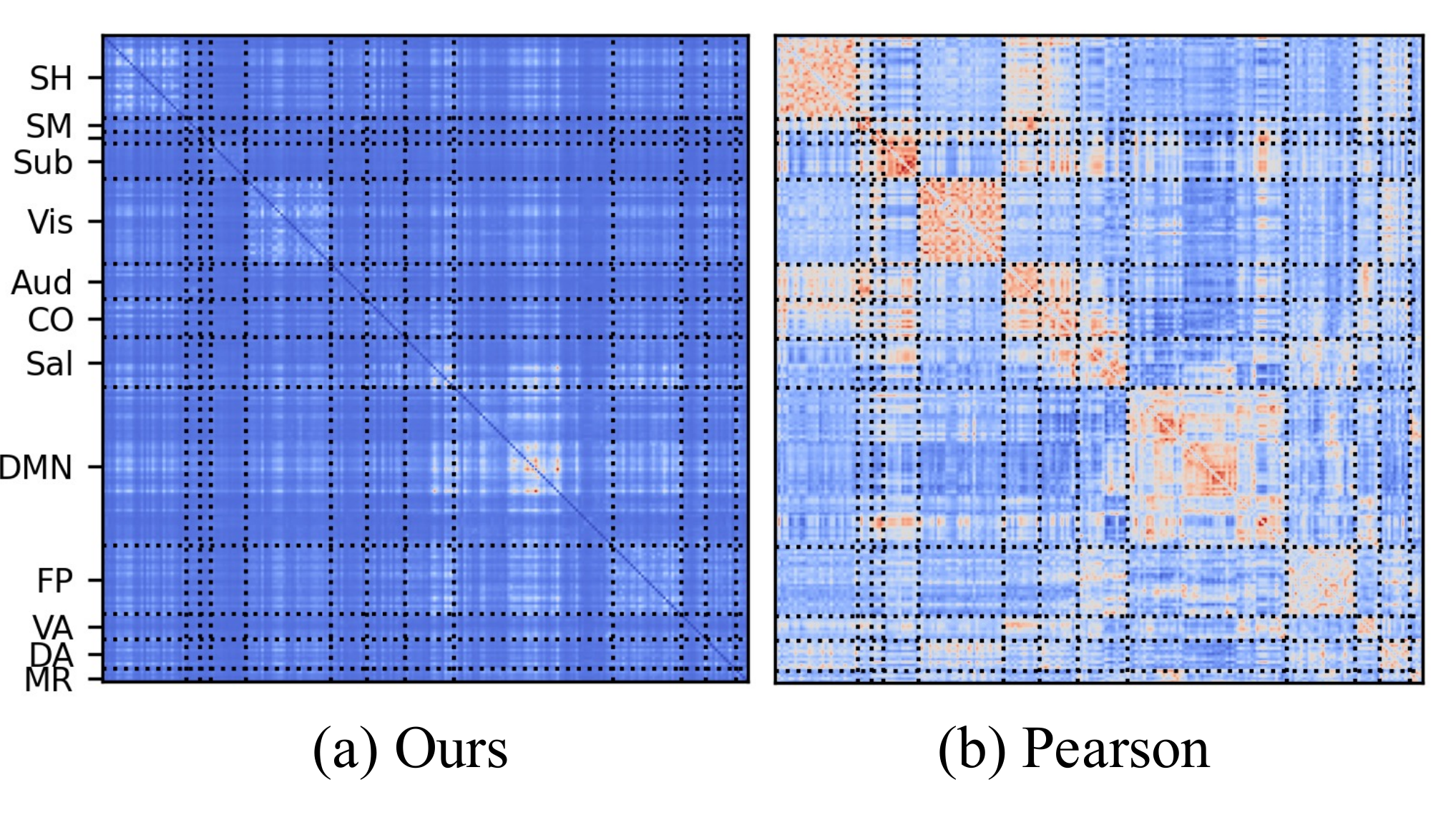}
    \caption{Visualizations of different brain networks. Abbreviations of neural systems: SH (Somatomotor Hand), SM (Somatomotor Mouth), Sub (Subcortical), Vis (Visual), Aud (Auditory), CO (Cingulo-opercular), Sal (Salience), DMN (Default mode), FP (Fronto-parietal), VA (Ventral attention), DA (Dorsal attention), MR (Memory retrieval). (a) is the mean heat map of all learnable brain graphs, while (b) is the mean heat map of all Pearson brain graphs.}
    \label{fig:explain}
\end{figure}
In this section, we visualize and compare our learnable graphs with the most commonly used existing functional brain networks, i.e., the Pearson Graphs \citep{modellingfmri}.
Results show that our learnable graph is more task-oriented and advantageous in capturing differences among classes. 

Specifically, we use the average graph of all samples to demonstrate the predominant neural systems consistent across subjects. Note that among all the 264 nodes in brain network, 32 of them cannot be grouped into any functional module. We remove them and the remaining 232 associated with the resting state networks are used for our connectivity analysis. 
The mean heat map visualization of our learnable brain graphs and Pearson brain graphs are shown in Figure \ref{fig:explain}. It is observed that our graph distinctively highlights the default mode network (DMN), while in the Pearson graph, the most significant positive components are the connections within functional modules. These within-module connections mainly reflect intrinsic brain functional organizations but are not necessarily informative for the gender prediction task. The highlight of DMN network in our learnable graph is consistent with the neurobiological findings on the PNC data \cite{genderfunction} that regions with significant differences between genders are located in the DMN. This comparison indicates that our learnable graphs can learn to highlight the prediction task-specific brain regions and connections, which are potentially useful for the analysis of other clinical prediction tasks where disease-region relevance is unclear. 

Besides, to demonstrate that our learnable graphs possess the discrimination ability among classes, in our experiments, we divide these learnable graphs based on genders. T-test is applied to identify edges $\mathcal{E}^{d}$ with significant difference ($p < 0.05$) between genders. Next, a difference score $T$ is calculated for each predefined functional module $u$,
$$
  T_u =   \sum_{(p,q) \in \mathcal{E}^{d}}\frac{\mathds{1}(p \in \mathcal{M}_u)+\mathds{1}(q \in \mathcal{M}_u)}{\abs{\mathcal{M}_u}}
$$
where $\mathcal{M}_u$ is a set containing all indexes of node belonging to the module $u$. Higher score indicates larger difference between genders. The top 3 modules are ventral attention network, default mode network and memory retrieval network. Our findings align well with results from previous work \citep{genderfunction} which show that most of the brain regions that demonstrate significant gender differences are located within these three modules. This  further validates that our learnable graphs can effectively capture group differences in brain networks, which facilitates imaging-based  classification.

\section{Conclusion}
In this paper, we present a framework which can generate the brain connectivity matrix and predict clinical outcomes simultaneously from fMRI BOLD signal series. 
Extensive experiments on our constructed learnable brain networks validate that the properly designed regularizers can help to unleash the power of GNNs and gain significant improvement over existing time series methods.
Besides, the visualization analysis of our learnable brain networks shows the generated networks is task-oriented and possess the discrimination ability among classes, providing a better interpretation towards tasks. 
In the near future, it would be interesting to explore the utility of our framework on more brain imaging datasets and downstream prediction tasks. 

\balance
\bibliographystyle{icml2021}
\bibliography{reference}

\end{document}